\documentclass[runningheads]{llncs}
\usepackage{graphicx}
\usepackage{amsmath}
\usepackage{amssymb}
\usepackage{multirow}
\usepackage{makecell}
\newcommand{\ie}{\textit{i}.\textit{e}.}
\newcommand{\eg}{\textit{e}.\textit{g}.}

\begin{document}
\title{Learning to Segment Anatomical Structures Accurately from One Exemplar}
%\titlerunning{Learning to Segment Anatomical Structures Accurately from One Exemplar}

\author{
Yuhang Lu\inst{1,2} \and
Weijian Li\inst{1,3} \and
Kang Zheng\inst{1} \and
Yirui Wang\inst{1} \and
Adam P. Harrison\inst{1} \and
Chihung Lin\inst{4} \and
Song Wang\inst{2} \and
Jing Xiao\inst{5} \and
Le Lu\inst{1} \and
Chang-Fu Kuo\inst{4} \and
Shun Miao\inst{1}
}
% index{Lu, Yuhang}
% index{Li, Weijian}
% index{Zheng, Kang}
% index{Wang, Yirui}
% index{Harrison, Adam}
% index{Lin, Chihung}
% index{Wang, Song}
% index{Xiao, Jing}
% index{Lu, Le}
% index{Kuo, Chang-Fu}
% index{Miao, Shun}

\authorrunning{Y. Lu et al.}
\institute{
PAII Inc., Bethesda, MD, USA \and
University of South Carolina, Columbia, SC, USA \and
University of Rochester, Rochester, NY, USA\and
Chang Gung Memorial Hospital, Linkou, Taiwan, ROC \and
Ping An Technology, Shenzhen, China}

\maketitle

\begin{abstract}
Accurate segmentation of critical anatomical structures is at the core of medical image analysis. The main bottleneck lies in gathering the requisite expert-labeled image annotations in a scalable manner. Methods that permit to produce accurate anatomical structure segmentation without using a large amount of fully annotated training images are highly desirable. In this work, we propose a novel contribution of \textit{Contour Transformer Network} (CTN), a one-shot anatomy segmentor including a naturally built-in human-in-the-loop mechanism. Segmentation is formulated by learning a contour evolution behavior process based on graph convolutional networks (GCNs). Training of our CTN model requires only one labeled image exemplar and leverages additional unlabeled data through newly introduced loss functions that measure the global shape and appearance consistency of contours. We demonstrate that our one-shot learning method significantly outperforms non-learning-based methods and performs competitively to the state-of-the-art fully supervised deep learning approaches. With minimal human-in-the-loop editing feedback, the segmentation performance can be further improved and tailored towards the observer desired outcomes. This can facilitate the clinician designed imaging-based biomarker assessments (to support personalized quantitative clinical diagnosis) and outperforms fully supervised baselines.

\end{abstract}
\section{Introduction}
Obtaining manual image segmentation labels or masks has often been an obstacle in scaling up medical image segmentation applications.
Without abundant pixel-level fully annotated image data, the state-of-the-art CNN-based segmentation methods cannot achieve their best performances~\cite{ronneberger2015u,dolz2018hyperdense,RothLLHFSS18,sinha2019multi,wang2019deep,chen2018encoder,harrison2017progressive}. However, annotating segmentation masks for medical images is very time-consuming and requires specialized expertise on human anatomy and its variations~\cite{tajbakhsh2019embracing}.
How to train an accurate segmentation model with less labeled data demands prompt solutions.
In this paper, we tackle the problem on high-resolution anatomical structure X-ray images and propose \textit{Contour Transformer Network}, allowing to learn from only one labeled exemplar image.

%In recent years, supervised deep learning approaches achieved the state-of-the-art performance in many medical image segmentation applications~\cite{ronneberger2015u,dolz2018hyperdense,RothLLHFSS18,sinha2019multi,wang2019deep,chen2018encoder,zhao2017pyramid,harrison2017progressive}. However, these methods are fully-supervised and require a large number of manual segmentation labels for training, which are usually difficult to obtain in practice~\cite{tajbakhsh2019embracing}. To reduce the dependence on manual labels, we study the problem of one-shot segmentation problem in the paper, and propose the Contour Transformer Network (CTN) for anatomical structure segmentation which could be trained with only one labeled image.

Several one-shot or few-shot segmentation methods have been proposed for natural images~\cite{shaban2017one,michaelis2018one,dong2018few,hu2018learning,zhang2019canet} by extracting information from a few support images to guide the segmentation of query images in testing. Nevertheless, the training process still relies on large-scale annotated datasets such as PASCAL VOC \cite{Everingham10} and MS-COCO \cite{LinMBHPRDZ14}. This condition renders them inapplicable directly to the medical imaging domain, because such equivalent datasets do not exist yet. Our problem is defined under a very different setting that only one pixel-level annotated training image instance is available. Other problem settings could be found in~\cite{oliveira2017augmenting,zhao2019data} where they attempt to alleviate the label shortage problem via data augmentation. In contrast, our method is to train the segmentation model with only one labeled exemplar and a set of unlabeled images.

The main challenge of one-shot segmentation is the lack of ground truth image mask or contour labels. Regular training strategies of comparing predictions with ground truth labels are no longer applicable. We adopt a new training scheme in CTN.
%Instead of directly comparing with the ground truth, we adopt the training strategy of making the prediction to be similar with the exemplar ``in some aspects''.
%Our training involves an exemplar image and its segmentation mask and a set of unlabeled images.
Because of the inherent regularized nature of anatomical structures, the same anatomy in different (X-ray) images may share some common features or properties, such as the anatomical structure's \textit{shape}, \textit{appearance} and \textit{gradients} along the structural object boundary. Although different images are not directly comparable, we can compare their common features only and use the exemplar segmentation to guide other unlabeled images partially, thus making CTN trainable in a one-shot setting.
% human behavior

To leverage these shared anatomical properties, we represent the image segmentation problem as learning a contour evolution behavior.
%We parameterize each contour in a sorted and uniformly manner to make them comparable.
Thus three differentiable contour-based loss functions are proposed to describe the common features. For each unlabeled image, CTN takes the exemplar contour as an initialization, then gradually evolves it to minimize the weighted loss. %The implementation of CTN takes inspirations from Curve-GCN's network architecture and classic Active Contour Models' energy terms.
Furthermore, we offer a naturally built-in human-in-the-loop mechanism to allow CTN to learn from extra partial labels. If any part in the predicted contour is inaccurate, users can correct them by drawing line segments, then CTN will format these corrections as partial contours and incorporate them back into the training via an additional Chamfer loss. In this way, we can improve and refine the segmentation performance with minimum annotation costs.
%Whenever CTN makes mistakes, the user could correct the wrong part of predicted contours, then CTN formats human corrections as partial contours and incorporated them back into the training via an additional Chamfer loss.
%In this way, we could achieve satisfactory performance with minimum annotation cost.
%This is an important feature because reducing annotation cost should not at the cost of performance.
%Namely, we format manual corrections as partial contours where users need only redraw incorrectly segmented parts, and leave correct parts untouched.
%These partial contour annotations can be naturally incorporated back into the training via an additional Chamfer loss.
%which means CTN could reach the performance of fully supervised methods but at lower annotation cost.

In summary, our contributions are three folds. (1) We propose a CNN-based image segmentation framework that could be trained with only one labeled image. (2) We describe the contour perceptual loss and the contour bending loss as two new optimization loss functions, to measure the similarity of two contours in terms of the appearance or shape cues, respectively. (3) We demonstrate that CTN achieves the state-of-the-art one-shot segmentation results; performs competitively when compared to fully supervised alternatives; and can outperform them with minimal human-in-the-loop feedback, on three datasets.

%-------------------------------------------------------------------------

\begin{figure*}[htbp]
	\begin{center}
		% \fbox{\rule{0pt}{2in} \rule{.9\linewidth}{0pt}}
		\includegraphics[width=\linewidth]{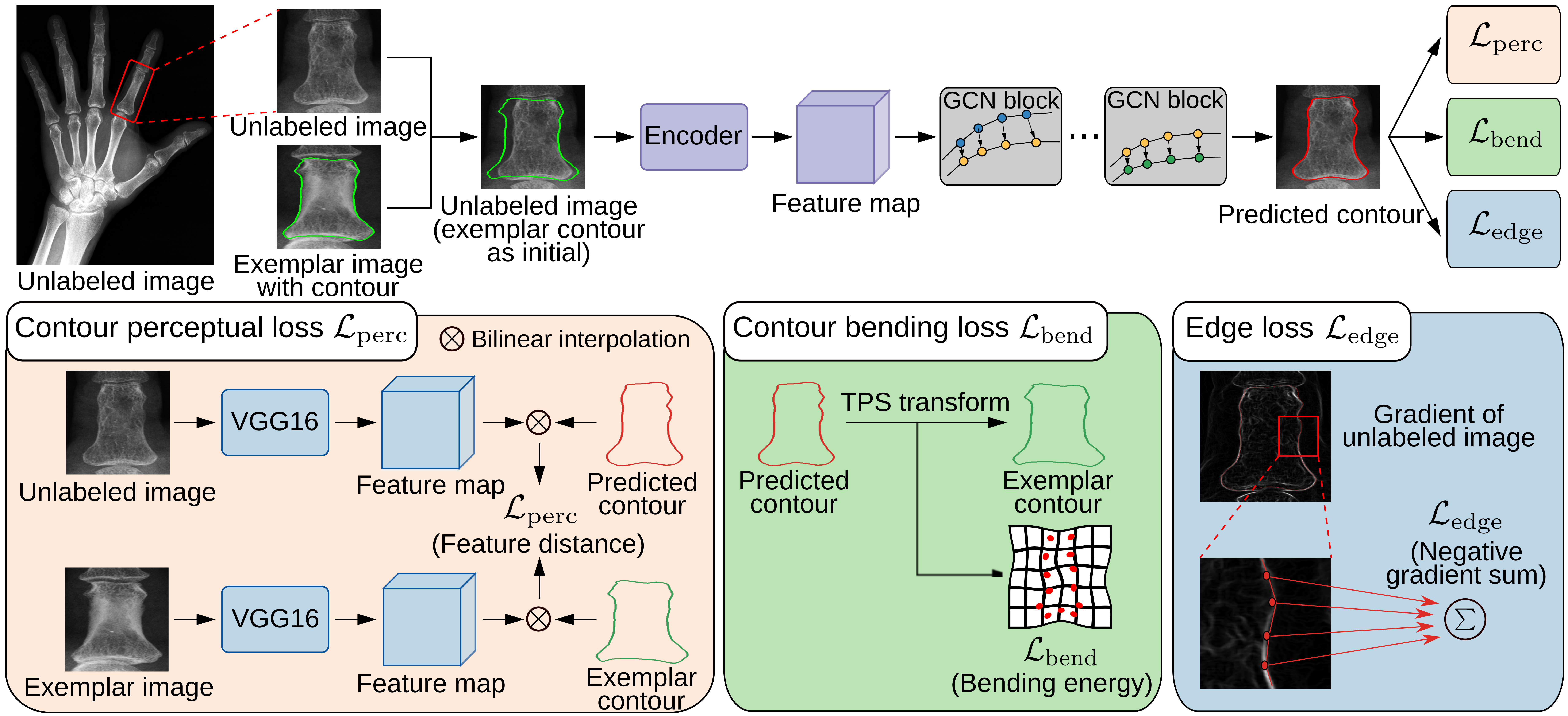}
	\end{center} %\vspace{-6mm}
	\caption{Contour Transformer Network. CTN is trained to fit a contour to the object boundary by learning from one labeled exemplar. It takes the exemplar and an unlabeled image as input, and predict a contour that has similar contour features with the exemplar. Three losses are proposed to make this network ``one-shot" trainable.}
	\label{fig:framework} %\vspace{-4mm}
\end{figure*}

\section{Methods}
The problem of anatomical structure segmentation on images can be decomposed into two steps: ROI (Region of Interest) cropping; and ROI segmentation. %Take phalanx segmentation on hand X-ray images as example, we first detect the bounding boxes of each phalanx and then perform image segmentation on each cropped phalanx ROI.
ROI detection has been well-studied in past literature~\cite{wu2018look,chen2019cephalometric,wang2019deep,wang2019weakly,li2020structured,chen2020anatomyaware}, so we focus on achieving very high segmentation accuracy by taking the detected/cropped ROI (with noise and errors) as input.

Assuming that a set of images $\mathbf{I}$ contains the same type of anatomical structure and only one of them is labeled, called the \textit{exemplar}. Our goal is to learn a segmentation model for this structure from $\mathbf{I}$. As mentioned above, we frame image segmentation as a process of contour evolution. Each contour is represented by $N$ uniformly spaced vertices. Denote the exemplar image and its contour by $I_E$ and $C_E$, respectively. For any unlabeled image $I \in \mathbf{I}$, its contour is $C=\{\mathbf{p}_1, \mathbf{p}_2, \dots, \mathbf{p}_N\}$. The exemplar contour is placed at the center of $I$ as the initial location of $C$;  next CTN is employed to estimate the point-wise offsets from the initial to the correct location, which is formulated by
$F_{\boldsymbol{\theta}}(I, I_E, C_E) = \{\Delta\mathbf{p}_1,\Delta\mathbf{p}_2,\dots,\Delta\mathbf{p}_N\}$
where $F_{\boldsymbol{\theta}}$ denotes the CTN model with weights $\boldsymbol{\theta}$.

%\noindent\textbf{Network Architecture.}
Inspired by~\cite{ling2019fast}, we use a \textit{CNN-GCN network architecture} to model the contour evolution in CTN. From Fig.~\ref{fig:framework}, CTN consists of two parts of an image encoding block and cascaded contour evolution blocks. It takes an unlabeled image $I$, an exemplar image $I_E$ and its ground truth contour $C_E$ as input, and predicts the contour $C$ of $I$. (1) We first place $C_E$ at the center of $I$ as the initial location of $C$, then the encoder outputs a feature map encoding the local image appearance of $I$. ResNet-50~\cite{he2016deep} is used as the backbone of CNN encoder. (2) The cascaded GCN blocks are then employed to evolve the contour $C$ step by step. The GCN takes the contour graph with vertex features as input. Each vertex in the contour is connected to four neighboring vertices, two on each side. These vertex features are extracted from the feature map of $I$ at the vertex locations via interpolation. Each GCN block takes the output contour of the previous block and updates it by predicting the point-wise coordinate offsets. We use five GCN blocks with the same multi-layer GCN architecture, although weights are not shared. The output of the 5\textit{th} block is the predicted contour of CTN. (3) Three one-shot trainable losses are utilized to optimize CTN, as the contour perceptual loss $L_{perc}$, the contour bending loss $L_{bend}$ and the edge loss $L_{edge}$. The total loss of CTN in the one-shot setting is written as:
$L=\lambda_1 L_{perc} + \lambda_2 L_{bend} + \lambda_3 L_{edge}$
where $\lambda_1, \lambda_2, \lambda_3$ are the weighting factors of the three losses. We describe the three employed losses in detail as follows.
%These losses imitate the human's behavior in learning contouring from one exemplar, i.e., drawing new contours by referring to the exemplar to compare shapes and local appearances. Another key insight is that although these losses can be used in an \ac{ACM} setting, where the contour vertices are directly optimized to minimize the energy, training \acp{our} on aggregating over the entire unlabeled dataset is robust, stable and can inhibit the boundary leaking issue on individual cases often encountered by \acp{ACM}.

%-------------------------------------------------------------------------
%\subsection{One-shot training losses} %\label{sec:loss}

%\subsubsection{Contour perceptual loss} %\label{sec:percep_loss}

% One training target of \ac{our} is to minimize
\noindent{\bf Contour Perceptual Loss.} We propose a new contour perceptual loss to measure the \textit{appearance dissimilarity} between the visual patterns of the exemplar contour $C_E$ and the predicted contour $C$, on the exemplar image $I_E$ or the target image $I$, respectively. Partially motivated by the original perceptual loss~\cite{johnson2016perceptual} developed for image super-resolution, modeling the image perceptual similarities in the feature space of VGG-Net~\cite{vgg}, we measure the contour perceptual similarities in the graph feature space. In particular, graph features are extracted from the ImageNet pre-trained VGG-16 feature maps of the two images along the two contours, and their L1 distance is calculated as the contour perceptual loss: $L_{perc} = \sum_{i=1,\dots,N} \| P(\mathbf{p}_i) - P_E(\mathbf{p}'_i) \|_1$
where $\mathbf{p}_i \in C$, $\mathbf{p}'_i \in C_E$, and $P$ and $P_E$ denote the VGG-16 features of $I$ and $I_E$, respectively.
The VGG-16 baseline network weights are trained on ImageNet dataset~\cite{imagenet}.

%Instead of using the L2 distance in the original perceptual loss~\cite{johnson2016perceptual}, we employ the L1 distance. Because there are always inevitable {\bf appearance variations} across images, we hypothesize that the similarity representation between any pairs of local image patterns is often limited according to certain aspects, \eg, specific texture, context, or shape features. Given that different channels of VGG-16 features capture different characteristics of local image patterns, a distance metric learning with modeling flexibility to select which salient features to match may be more appropriate. The sparsity-inducing nature of L1 distance definition provides the additional ``selection'' mechanism over L2. It also indeed empirically performs better in our experiments. %which may explain the improved performance observed.

The contour perceptual loss is used to guide the evolution of the contour in CTN, by having several advantages. (1) Since VGG-16 network features can capture the image pattern of a neighboring area with spatial contexts (\ie, network receptive field), the contour perceptual loss enjoys a relatively large capturing range (\ie, the convex region around the minimum), making the CTN training optimization easier. (2) The backbone VGG-16 model is trained on ImageNet ~\cite{imagenet} for classification tasks, so that its learned features are less sensitive to noises and illumination variations, which also benefits the training of CTN.

\noindent{\bf Contour Bending Loss.} If we operate under the assumption that an exemplar contour is broadly informative to other data samples, then it should be beneficial to use the exemplar shape to ground any predictions on other samples. To this end, we propose a novel contour bending loss to measure the \textit{shape dissimilarity} between contours. The loss is calculated as the bending energy of thin-plate spline (TPS) warping~\cite{bookstein1989principal} that maps $C_E$ to $C$. It is worth noting that TPS warping achieves the minimum bending energy among all warpings that map $C_E$ to $C$.
Since the bending energy measures the magnitude of the second order derivatives of the warping function, the contour bending loss penalizes more on the local and acute shape changes, often associated with mis-segmentations. Given $C$ and $C_E$, the TPS bending energy can be calculated as follows. Define $\mathbf{p}_i=(x_i, y_i)$, $\mathbf{p}'_i=(x'_i, y'_i)$, and~
$\mathbf{K} = \left ( \left \| \mathbf{p}^{\prime}_i - \mathbf{p}^{\prime}_j \right \|_2 ^ 2 \cdot log\left \| \mathbf{p}^{\prime}_i - \mathbf{p}^{\prime}_j \right \|_2 \right )$, $    \mathbf{P} = (\mathbf{1}, \mathbf{x}^{\prime}, \mathbf{y}^{\prime})$, $
    \mathbf{L} = \begin{bmatrix} \mathbf{K} & \mathbf{P}\\ \mathbf{P}^T & \mathbf{0}\end{bmatrix}$
where $\mathbf{x}^{\prime}=\{x^{\prime}_1, x^{\prime}_2, \ldots, x^{\prime}_N\}^T$,  $\mathbf{y}^{\prime}=\{y^{\prime}_1, y^{\prime}_2, \ldots, y^{\prime}_N\}^T$.
The TPS bending energy is written as
$ L_{bend} = \max \left [ \frac{1}{8\pi}(\mathbf{x}^T \mathbf{H} \mathbf{x} + \mathbf{y}^T \mathbf{H} \mathbf{y}) ,0 \right ]$
where $\mathbf{x}=\{x_1, x_2, \ldots, x_N\}^T$, $\mathbf{y}=\{y_1, y_2, \ldots, y_N\}^T$, and $\mathbf{H}$ is the $N \times N$ upper left submatrix of $\mathbf{L}^{-1}$~\cite{wang2017correspondence}.

\noindent{\bf Edge Loss.} Although the contour perceptual and bending losses can achieve robust segmentation, they are inherently insensitive to (very) small segmentation fluctuations, such as minimal deviations from the correct boundary by a few pixels. Therefore, in order to obtain desirably high segmentation accuracy and adequately facilitate the downstream workflows like rheumatoid arthritis quantification~\cite{huo2015automatic}, we employ an edge loss that measures the image gradient magnitude along the computed contour and attracts the contour toward edges in the image naturally. The edge loss is written as:
$L_{edge} = - \frac{1}{N} \sum_{\mathbf{p} \in C} {\left \| \nabla I(\mathbf{p}) \right \|_2}$ where $\nabla$ is the gradient operator.

\subsection{Human-in-the-loop}
%\noindent{\bf Human-in-the-loop.}
%Reducing the annotation cost should not be at the cost of performance on accuracy.
More labels are always helpful to enhance the model's generalization ability and robustness, if available.
Benefiting from the contour-based setting, CTN offers a natural way to incorporate additional user labels with a human-in-the-loop mechanism. Assuming that we have a CTN model trained with one exemplar image, we intend to finetune it with more segmentation annotations. We run this model on a set of unlabeled images first, and select any number of images with inaccurate predictions as new instances. Instead of drawing the whole contour from scratch on these new images, the annotator only needs to redraw some partial contours, to correct the previously undesirable predictions. The point-wise training of CTN makes it feasible to learn from these partial corrections. %We reduce the annotation labor cost to the minimum.

A \textbf{partial contour matching loss} is proposed to utilize the partial ground truth contours during the CTN training. Denote $\hat{\mathbf{C}}$ as a set of partial contours in image $I$, each element of which is an individual contour segment. For each contour segment $\hat{C}_i \in \hat{\mathbf{C}}$, we build the point correspondence between $\hat{C}_i$ and $C$. For each $\hat{C}_i$, we find two points in the predicted contour $C$, closest to the start and end points of $\hat{C}_i$, then each predicted point between the two points are assigned to the closest corrected point. Denote the corresponding predicted contour segment by $C_i$ ($C_i \in C$). We define the distance between $C$ and $\hat{C}_i$ as the Chamfer distance from $C_i$ to $\hat{C}_i$:
$ D(\hat{C}_i, C) = \sum_{\mathbf{p} \in C_i}\min_{\mathbf{\hat{p}} \in \hat{C}_i}\left \| \mathbf{p}- \mathbf{\hat{p}} \right \|_2$ and the partial matching loss of $C$ is defined as
$L_{pcm} = \frac{1}{N}\sum_{\hat{C}_i \in \hat{\mathbf{C}}} D(\hat{C}_i, C).$ In the human-in-the-loop scenario, we combine all losses to train the CTN, and rewrite the loss function as
$\hat{L}=\lambda_1  L_{perc} + \lambda_2 L_{bend} + \lambda_3 L_{edge} + \lambda_4 L_{pcm}$,
which allows CTN to be trained with fully labeled, partially labeled and unlabeled images simultaneously and seamlessly. Whenever new labels are available, we can use $\hat{L}$ to finetune the one-shot CTN model.

\begin{table*}[htbp]
\caption{Performances of CTN and seven existing methods on three datasets.}
\begin{center}%\vspace{-6mm}
\setlength{\tabcolsep}{0.9mm}
\begin{tabular}{c|c|c c|c c|c c}
\hline
\multicolumn{2}{c|}{\multirow{2}{*}{Method}}                    & \multicolumn{2}{c|}{Knee}                                & \multicolumn{2}{c|}{Lung}                                & \multicolumn{2}{c}{Phalanx} \\ \cline{3-8}
\multicolumn{2}{c|}{}                                           & IoU(\%)                          & HD(px)                        & IoU(\%)                          & HD(px)                        & IoU(\%)           & HD(px)          \\ \hline
\hline
\multirow{2}{*}{\makecell{Non- \\ learning}}   & MorphACWE~\cite{caselles1997geodesic,marquez2013morphological}                       & 65.89                      & 54.07                     & 76.09                      & 55.35                     & 74.33       & 69.13      \\
                              & MorphGAC~\cite{chan2001active,marquez2013morphological}                        & 87.42                      & 15.78                     & 70.79                      & 45.67                     & 82.15       & 24.73      \\ \hline
\hline
\multirow{3}{*}{\makecell{One- \\ shot}}     & CANet~\cite{zhang2019canet}                           & 29.22                      & 175.86                    & 56.90                      & 73.46                     & 60.90       & 67.13      \\
                              & Brainstorm~\cite{zhao2019data}                      & 90.17                      & 29.07                     & 77.13                      &43.28                      & 80.05       & 30.30        \\
                              & \textbf{CTN (Ours)}                      & \textbf{97.32}             & \textbf{6.01}             & \textbf{94.75}             & \textbf{12.16}            & \textbf{96.96} & \textbf{8.19}       \\ \hline
\hline
\multirow{3}{*}{\makecell{Fully \\ supervised}}   & UNet~\cite{ronneberger2015u}                            & 96.60                      & 7.14                      & 95.38                      & 12.48                     & 95.76       & 10.10      \\
                              & DeepLab~\cite{chen2018encoder}                     & 97.18                      & 5.41                      & 96.18                      & 10.81                     & 97.63       & 6.52       \\
                              & HRNet~\cite{wang2019deep}                           & 96.99                      & 5.18                      & 95.99                      & 10.44                     & 97.47       & 7.03       \\ \hline
\end{tabular}
\end{center}
\label{table:compare} %\vspace{-9mm}
\end{table*}

\section{Experimental Results}

%{\bf Datasets and Experimental Settings}
\noindent\textbf{Datasets.} We evaluate our method on three X-ray image datasets of knee, lung and phalanx, respectively. The \textit{knee} dataset contains 212 knee X-ray images from the Osteoarthritis Initiative (OAI) database\footnote{\url{https://nda.nih.gov/oai/}}, 100 for training and 112 for testing. The \textit{lung} dataset is the public JSRT~\cite{JSRTdatabase} of 247 posterior-anterior chest radiographs, 124 for training and 123 for testing. The \textit{phalanx} dataset comes from hand X-ray images of patients with rheumatoid arthritis. 202 ROIs of proximal phalanx are extracted automatically from hand joint  detection~\cite{huo2015automatic}, randomly split into 100 training and 102 testing images.

%\noindent\textbf{Evaluation metrics.}
We evaluate the accuracy of segmentation masks by Intersection-over-Union (IoU) metric and the corresponding object contour distance by the Hausdorff distance (HD). For comparative methods not explicitly outputting the anatomy contours, we extract the external contour of the largest segmented image region. The hyperparameters are $N = 1000$, $\lambda_1 = 1$, $\lambda_2 = 0.25$, $\lambda_3 = 0.1$, $\lambda_4 = 1$. All networks are trained using Adam optimizer with a learning rate of $1\times10^{-4}$, a weight decay of $1\times10^{-4}$ and a batch size of 12 for 500 epochs. The one-shot training and human-in-the-loop finetuning settings are the same.

%Two evaluation metrics are employed in our experiment, the Intersection-over-Union (IoU) and the Hausdorff Distance (HD). The mean IoU of all foreground classes is commonly used in segmentation evaluation. But it focuses on the full region and is not sensitive to minor errors on the object boundary~\cite{krahenbuhl2014geodesic}. Thus we also report the Hausdorff Distance from the predicted contour to the ground truth object boundary to evaluate the accuracy along the predicted contour. For those heatmap-based methods that do not explicitly output contour, we use the external contour of the largest region of each class in their predicted masks.

%-------------------------------------------------------------------------

%\noindent\textbf{Implementation details.}

% add how to select exemplar

%-------------------------------------------------------------------------

\begin{figure*}[htbp]
% \vspace{-1em}
	\begin{center}
%		\fbox{\rule{0pt}{2in} \rule{.9\linewidth}{0pt}}
		\includegraphics[width=\linewidth]{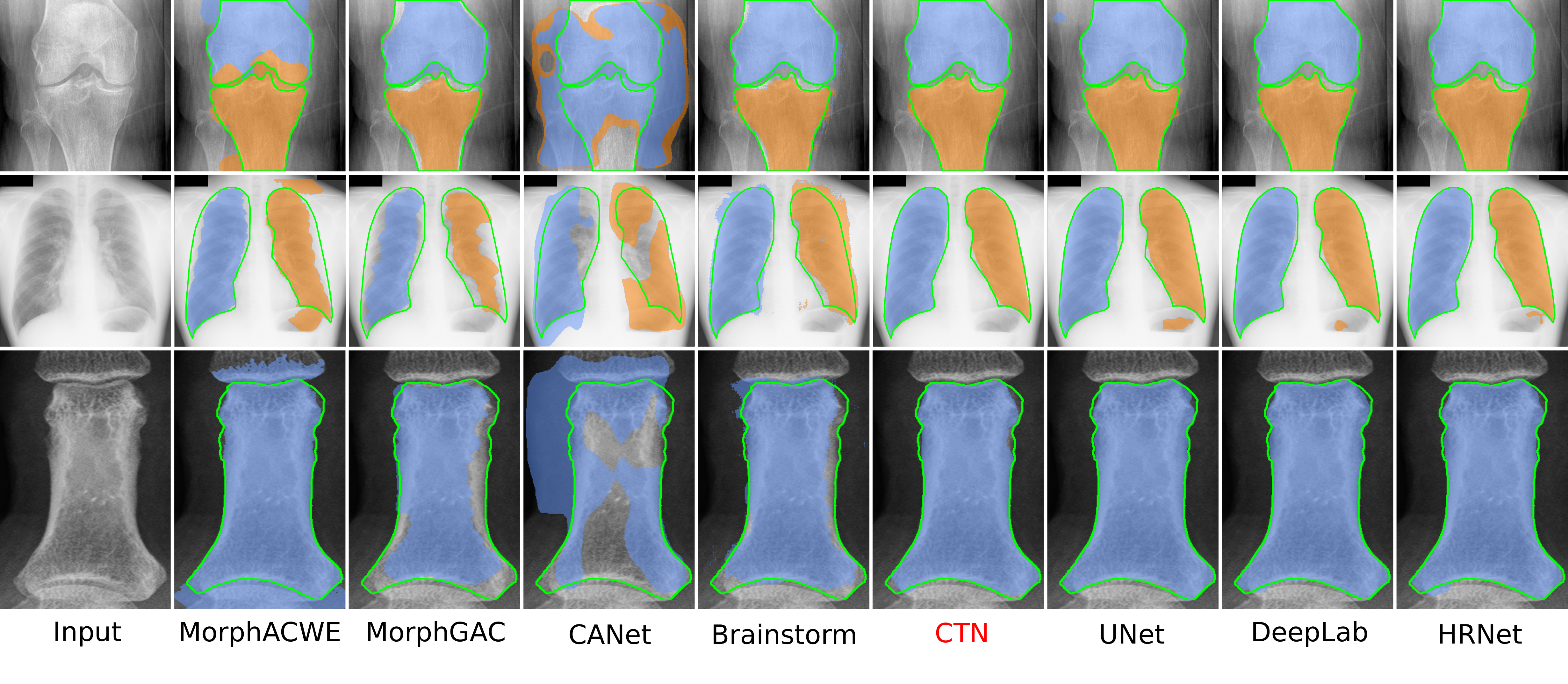}
	\end{center} %\vspace{-9mm}
	\caption{Segmentation results of three example images using eight methods. From top to bottom, these images are from the knee, lung and phalanx testing sets, respectively. The ground truth boundaries are drawn in \textbf{green} line for the ease of comparison.}
	\label{fig:seg_compare} %\vspace{-7mm}
\end{figure*}

The proposed CTN is compared with seven previous methods. The quantitative results are reported in Table~\ref{table:compare}; qualitative results are given in Fig.~\ref{fig:seg_compare}.

\noindent\textbf{Comparison with non-learning-based methods.} We first compare with two non-learning based methods of MorphACWE~\cite{caselles1997geodesic,marquez2013morphological} and MorphGAC~\cite{chan2001active,marquez2013morphological}. Both approaches are based on active contour models (ACMs), which evolves an initial contour to the object by minimizing an energy function. The initial contour is the same as ours. Quantitative results in Table~\ref{table:compare} show that our method significantly outperforms them. Specifically, in average CTN achieves 16.22\% higher IoU and 19.94 pixels less in HD than MorphGAC, the better of the two. Visualization results in Fig.~\ref{fig:seg_compare} show that these two approaches cannot localize anatomical structures accurately, especially when the boundary of such structures are not clearly contrasted, such as in the lung image. Both methods are based on ACMs and predict contours by minimizing some hand-crafted energy functions for a \textit{single} image. In contrast, CTN learns from an exemplar contour to guide the contour transformation for all images in the entire training set.

%The inferior performances of both MorphACWE and MorphGAC result from their hand-crafted features and sample-wise energy minimization formulations, which are not robust to image noise or other possible variations.

% Unlike non-learning-based methods that use handcrafted energy functions to guide the contour evolution, CTN utilizes an exemplar contour to do so, which is more intuitive. Besides, non-learning-based methods minimizes the energy on a single image, while CTN minimizes the loss on the entire training set.

%The results shown in Table.~\ref{table:compare} shows that our method outperforms MorphGAC, the better of the two non-learning-based methods, by a margin of 16.14\% in terms of average IoU. \textcolor{red}{Both MorphACWE and MorphGAC are based on active contour models, which minimizes certain energy functions such that the contours are attracted to edges in images. However, these methods are sensitive to image noise and may produce leakages. For example, as shown in the third column Fig.~\ref{fig:seg_compare}, MorphGAC performs reasonably on the knee image, but shows large leakage in both lung and phalanx.}

\noindent\textbf{Comparison with other one-shot methods.} Next, we compare with two state-of-the-art one-shot segmentation methods: CANet~\cite{zhang2019canet} and Brainstorm~\cite{zhao2019data}. All one-shot approaches (including ours) use the same exemplar image. For each training set, we compare the distance of each image to all other images in the VGG network feature space and the exemplar is selected to be the image with the smallest distance. CANet is proposed to perform one-shot segmentation on unseen objects in testing, but it requires a fully annotated dataset to train. Such that, we use the specific model~\cite{zhang2019canet} trained on PASCAL VOC 2012 dataset for comparison. From Table~\ref{table:compare}, CANet achieves only $49.01\%$ IoU on average. We speculate that the poor performance is caused by the domain gap between natural images and medical images. Brainstorm~\cite{zhao2019data} learns an image augmentation model and is one-shot trainable. We follow its default procedure to train the segmentation models on three datasets. It yields reasonable results with averaged $82.45\%$ IoU and $34.22$ HD, but still dramatically lower than ours.

\noindent\textbf{Comparison with fully supervised methods.} Last, we evaluate and compare the performance of three fully supervised methods: UNet~\cite{ronneberger2015u}, DeepLab v3+~\cite{chen2018encoder} and HRNet W18~\cite{wang2019deep}. All of them are trained for 500 epochs using all training image annotations in three datasets. On average, CTN performs comparably or better than UNet, and falls behind DeepLab (the best of the three), by $0.66\%$ in IoU and $1.21$ pixels in HD, respectively. It demonstrates that CTN while using only one training image can usually compete head-to-head with the state-of-the-art fully supervised baselines \cite{ronneberger2015u,chen2018encoder,wang2019deep}. Note that the heatmap-based segmentation methods predict the per-pixel labels, which could cause the loss of integrity of object boundaries, \eg, some small ``islands'' in the lung masks of Fig.~\ref{fig:seg_compare}. On the other hand, CTN naturally retains the integrity of the object segmentation, as an important aspect in assessing visual segmentation quality.

\begin{figure}[t]
    \begin{center}
    	\includegraphics[width=\linewidth]{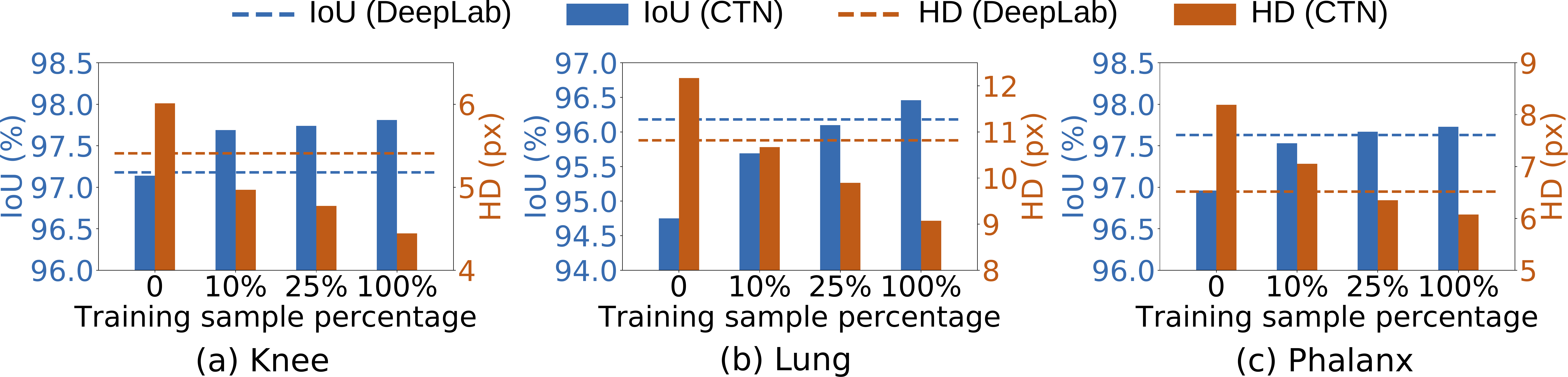}
    \end{center} %\vspace{-7mm}
    \caption{Using 0, 10\%, 25\% and 100\% human corrections to finetune the one-shot CTN model, respectively (``0'' means no finetuning).}
    \label{fig:more_label} %\vspace{-5mm}
\end{figure}

\noindent\textbf{Incorporating simulated human corrections.} To evaluate the effectiveness of the human-in-the-loop mechanism, we empirically simulate different degrees of human-computer interactions. For each dataset, we first train a CTN model with the default exemplar and run inference on the training set. We sort all training images by their HD segmentation errors from high to low. Three subsets are formed by selecting the top 10\%, 25\% or 100\% training images; and fine-tune the initial one-shot CTN model using these training subsets augmented by the ground truth contours, respectively. %We show the performances of the original model (no human correction) and three fine-tuned models.
This protocol results in four CTN models. From Fig.~\ref{fig:more_label}, we observe that CTN consistently improves with more human corrections. Specifically, when using 25\% such corrected samples, CTN starts to outperform DeepLab using all training images (IoUs of 97.17\% vs 97.0\%, and HDs of 7.01 vs 7.58). With all samples, CTN reaches 97.33\% on IoU and 6.5 on HD. These results indicate that the human-in-the-loop mechanism can potentially help CTN achieves better performance than fully supervised methods with considerably less annotation efforts.

\noindent\textbf{Ablation study.} We conduct an ablation experiment to validate the effectiveness of the three proposed losses. The results are shown in Table~\ref{table:ablation} where the performance of our method is indeed impaired if any loss is removed, with mean IoU reductions of 4.32\%, 4.12\%, and 1.72\% for $L_{perc}$, $L_{bend}$, and $L_{edge}$, respectively. This validates the necessity of all three losses. An exception is the knee dataset when $L_{bend}$ is removed. Knee X-ray images share similar appearance features along the contour so that they can be segmented robustly with just the contour perceptual loss and edge loss. Thus, adding contour bending loss leads to statistically insignificant decreases (\ie, IoUs of 97.32\% vs 97.50\%, HDs of 6.01 vs 5.87) in this particular scenario. However, such a regularization effect by the contour bending loss is generally desired to alleviate the worst-case scenarios and proves useful in the other two datasets.

\setlength{\tabcolsep}{2mm}
\begin{table*}[ht] %\vspace{-4mm}
\caption{Ablation study. Remove one loss each time and re-train the model.}
\label{table:ablation}
\begin{center}  %\vspace{-9mm}
\begin{tabular}{ccc|cc|cc|cc}
\hline
\multirow{2}{*}{$L_{perc}$} & \multirow{2}{*}{$L_{bend}$} & \multirow{2}{*}{$L_{edge}$} & \multicolumn{2}{c|}{Knee} & \multicolumn{2}{c|}{Lung} & \multicolumn{2}{c}{Phalanx} \\ \cline{4-9}
                            &                              &                             & IoU(\%)          & HD(px)         & IoU(\%)          & HD(px)         & IoU(\%)           & HD(px)          \\ \hline
                            & $\checkmark$                 & $\checkmark$                & 94.62      & 8.28     & 87.45      & 26.51     & 94.01       & 15.81       \\
$\checkmark$                &                              & $\checkmark$                & 97.50      & 5.87     & 84.93      & 36.74     & 94.24       & 26.13      \\
$\checkmark$                & $\checkmark$                 &                             & 94.43      & 11.90      & 92.99      & 16.22      & 96.45       & 9.84      \\ \hline
$\checkmark$                & $\checkmark$                 & $\checkmark$                & 97.32       & 6.01      & 94.75       & 12.17      & 96.96        & 8.19       \\ \hline
\end{tabular}
\end{center} %\vspace{-12mm}
\end{table*}

\section{Conclusion}
In this paper, we propose a novel one-shot segmentation method, Contour Transformer Network, which takes one labeled exemplar and a set of unlabeled images to train a segmentation model for anatomical structures in medical images. The key idea that enables one-shot training is to guide the segmentation of unlabeled images by utilizing their shared features with the exemplar image but not ground truth masks.
%It is implemented with a contour evolution network and three contour-based one-shot losses.
%We also propose a human-in-the-loop mechanism to let CTN exploit additional labels.
Experiments on three datasets demonstrate that CTN performs competitively to state-of-the-art fully supervised approaches and outperforms them with minimal human corrections. Although CTN is for anatomical structures, the idea of one-shot training is also applicable to other images with shared features. In the future, we will explore its application in more medical image analysis problems.
% Experiment results on three datasets show that our method significantly outperforms non-learning-based methods and other one-shot methods, and is comparable with fully-supervised methods. By further introducing a human-in-the-loop mechanism, our method could beat fully-supervised methods while using 25\% labels of them on this task. In short, our method uses as few labels as possible to to achieve outstanding segmentation performance.


\bibliographystyle{splncs04}
\bibliography{ctn}

\end{document}